\documentclass[10pt,twocolumn,letterpaper]{article}
\usepackage[T1]{fontenc}
\usepackage[utf8]{inputenc}
\usepackage[letterpaper,textwidth=6.875in,textheight=8.875in,centering]{geometry}
\usepackage{times,xspace,graphicx,amsmath,amssymb,booktabs,tabularx,etoolbox}
\usepackage[dvipsnames]{xcolor}
\usepackage[numbers,sort&compress]{natbib}
\usepackage[format=plain,labelformat=simple,labelsep=period,font=small]{caption}
\usepackage[font=footnotesize,skip=3pt,subrefformat=parens]{subcaption}
\usepackage[hyphens]{url}
\usepackage[shortlabels,inline]{enumitem}
\definecolor{linkblue}{rgb}{0.21,0.49,0.74}
\usepackage[breaklinks,colorlinks,allcolors=linkblue]{hyperref}
\hypersetup{pdftitle={Zetta-RSR: Scaling VLA Post-Training with Interface-Guided Real-Sim-Real Data Collection}}
\title{Zetta-RSR: Scaling VLA Post-Training with Interface-Guided Real--Sim--Real Data Collection}
\usepackage{air-heading}
\author{%
  \airauthorfont
  Yikai Zheng\textsuperscript{1,*}\quad
  Fu Chen\textsuperscript{2,*}\quad
  Xin Ding\textsuperscript{2,*,\dag}\quad
  Kejun Gao\textsuperscript{1}\quad
  Xuanyi Xie\textsuperscript{1}\quad
  Xiangyu Li\textsuperscript{1}\quad
  Bingjia Huang\textsuperscript{3}\\[3pt]
  Kun Li\textsuperscript{1}\quad
  Hao Wu\textsuperscript{4}\quad
  Yunxin Liu\textsuperscript{1}\quad
  Ting Cao\textsuperscript{1,\dag}\\[5pt]
  \airaffilfont\textsuperscript{1}Institute for AI Industry Research (AIR), Tsinghua University\\
  \airaffilfont\textsuperscript{2}University of Science and Technology of China\\
  \airaffilfont\textsuperscript{3}University of Electronic Science and Technology of China\\[3pt]
  \airaffilfont\textsuperscript{4}Nanjing University\\[4pt]
  \airaffilfont\textsuperscript{*}Equal contribution.\quad
  \textsuperscript{\dag}Corresponding authors: Xin Ding, Ting Cao\\[2pt]
  \airaffilfont\href{mailto:xinding64@mail.ustc.edu.cn}{\texttt{xinding64@mail.ustc.edu.cn}}\quad
  \href{mailto:tingcao@mail.tsinghua.edu.cn}{\texttt{tingcao@mail.tsinghua.edu.cn}}\\[4pt]
  \airaffilfont Project page: \url{https://air-embodied-brain.github.io/zetta-rsr/}%
}

\date{}

\makeatletter
\apptocmd{\@maketitle}{% Non-floating first-page teaser, appended to the full-width title block.
\begin{minipage}{\textwidth}
  \centering
  \captionsetup{hypcap=false}
  \includegraphics[width=\textwidth]{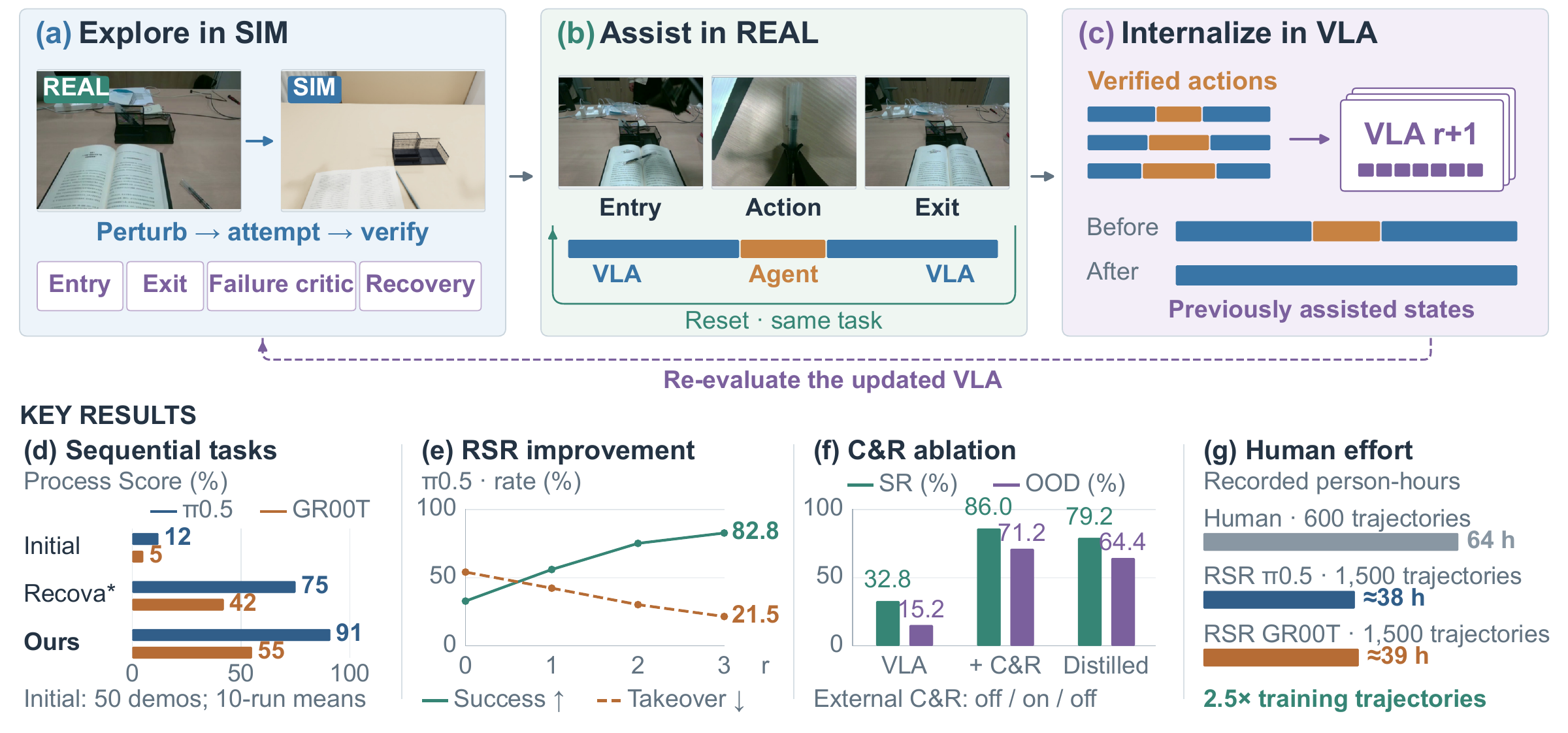}
  \captionof{figure}{\textbf{Zetta-RSR: from agent assistance to autonomous action capability.}
  (a) Simulation probes the current VLA and develops skill interfaces.
  (b) The agent assists real-robot execution, with Reset enabling repeated collection for the same task.
  (c) Verified action trajectories support VLA post-training and the next RSR round.
  (d) Ten-run Process Scores for both backbones (initial VLA: 50 demonstrations; Recova*: mechanism adaptation).
  (e) $\pi_{0.5}$ success and agent takeover across RSR rounds.
  (f) C\&R ablation, with external C\&R enabled only for the assisted original VLA and disabled for the distilled VLA.
  (g) Recorded human effort, including initial collection, for 600 human-only versus 1,500 mixed-source training trajectories.
  \emph{Source images and control ribbons illustrate the mechanism; panels (d--g) show reported experimental results.}}
  \label{fig:teaser}
\end{minipage}
\par\vspace{8pt}
}{}{\PackageError{zetta-rsr}{Could not attach the teaser}{}}
\makeatother

\begin{document}
\maketitle
% English translation of the approved concise Chinese abstract, 2026-11-16.
% Numerical results remain placeholders pending experimental verification.
\begin{abstract}
Continual post-training of vision-language-action (VLA) models is constrained by the cost of human data collection. We propose Zetta-RSR, which bootstraps multiple rounds of autonomous data collection and training through Real--Sim--Real from a small set of human demonstrations. In simulation, the agent explicitly explores the VLA's capability boundary and constructs skill interfaces grounded in states observable across domains to guide skill handoff and failure recovery on the real robot. Verified real-world corrective trajectories are used to post-train the VLA, and its updated capability boundary guides exploration in the next round, progressively transforming the agent's external assistance into the policy's autonomous action capabilities. In evaluations on an ARX Lift robot with $\pi_{0.5}$ and GR00T, Zetta-RSR uses \textbf{6} hours of additional human effort to collect valid trajectories equivalent to \textbf{48} hours of human data collection, and improves OOD success rate by \textbf{51.6} percentage points while reducing agent takeover rate by \textbf{24.3} percentage points through multiple iterations.
\end{abstract}

% English translation of the approved Chinese Introduction, 2026-11-13.
% Five paragraphs correspond to introduction_long_horizon_zh.md.
\section{Introduction}
\label{sec:introduction}

The post-training of vision-language-action (VLA) models relies on interaction data that cover the target task and variations in its conditions to improve generalization to changes in object positions, visual conditions, and contact states~\cite{lin2025data, world2025agibot, pan2026sop}. Expanding these data through human teleoperation requires operators to remain involved in task execution, failure handling, and scene preparation, so increases in data volume and condition coverage entail higher human data collection costs~\cite{khazatsky2024droid, mandlekar2020human, kelly2019hg}. These costs constrain the continued expansion of post-training data. We focus on autonomous post-training bootstrapped from a small set of human demonstrations, in which robots, after acquiring basic task competence, autonomously collect new interaction experience for model updates, progressively improving generalization and reducing dependence on human effort for further data expansion~\cite{fang2026find}.

Agent--VLA collaboration provides a foundation for reducing the resulting human intervention: the VLA is responsible for contact-intensive action execution, while the agent uses high-level reasoning, monitoring, and recovery to address states that the VLA cannot yet handle~\cite{zhang2026harness, ding2026zetta}; works such as Recova have used real task and recovery experience to update policies~\cite{liu2026recova}. However, such autonomous collection is particularly difficult in long-horizon tasks: a task consists of multiple interdependent operational stages, and the grasp poses, object positions, and contact relations left by earlier operations change the execution conditions of subsequent stages~\cite{hu2026rac}. For assisted execution to continuously expand the VLA's autonomous range, the system must also explicitly evaluate the current VLA's capability boundary, identify which states can proceed and which require agent takeover, and turn verified recovery trajectories into training supervision for these capability gaps.

A small number of human demonstrations can typically cover only limited operating conditions, whereas the intermediate states and failure cases caused by preceding deviations in long-horizon tasks are precisely capability gaps that the VLA still needs to address~\cite{hu2026rac}. Simulation provides a space for repeatedly exploring these states, but the reliable execution range identified there can change because of differences in observation and contact conditions on the real robot~\cite{li2026arstag,peng2018sim,wang2026reconstruct}, and can also change as the VLA is post-trained. Therefore, autonomous post-training needs to connect cross-domain-verifiable capability descriptions, agent assistance, and real-robot supervision, so that subsequent collection remains focused on the current model's actual capability gaps.

To this end, we propose Zetta-RSR (Figure~\ref{fig:teaser}), which starts from an initial VLA trained with a small number of human demonstrations and organizes multi-round Real--Sim--Real autonomous post-training. Simulation exploration centers on operational states in which the current model still requires agent assistance, while real interaction verifies the learned collaboration strategy under a limited adaptation budget and converts the corresponding capability judgments and actual corrections into training supervision for the VLA. After the model is updated, states that previously required assistance are reevaluated; the cases that have been mastered and those that still require takeover are used to revise the collaboration and collection objectives for the next round. This cycle aims to progressively internalize the agent's judgment and corrective capabilities into the policy's autonomous capabilities, expanding the range of states it can handle independently in long-horizon tasks.

This cycle is supported by a learnable skill interface~\cite{ju2026embodiskill}. Using conditions observable in both domains, such as relative object poses and grasp states, the interface models the current VLA's capability at each operational stage and the evidence supporting that estimate. Based on this capability description, the interface determines the VLA's entry, exit, and takeover conditions, coordinates the terminal states and entry requirements of adjacent stages, and specifies target states that can be restaged after failure recovery. The reasons for takeover, states before and after recovery, and actual corrective actions are thereby aligned, providing concrete supervision for the agent to when to intervene, and how to repair. 

% Approved experimental-summary wording; all numerical results remain placeholders.
We evaluate Zetta-RSR on an ARX Lift robot using two VLA backbones, $\pi_{0.5}$~\cite{intelligence2025pi_} and GR00T~\cite{bjorck2025gr00t}, across ten everyday manipulation tasks and continuous task execution in a room. After bootstrapping from a small set of human demonstrations, the system autonomously collects valid real-robot trajectories equivalent to \textbf{48} hours of human data collection, requiring only \textbf{6} hours of additional human effort. Across multiple RSR iterations, the average system success rate and OOD success rate on five representative tasks improve by \textbf{49.2} and \textbf{42.8} percentage points, respectively, while the agent takeover rate decreases from \textbf{57.7\%} to \textbf{33.4\%}, coupling performance gains with reduced dependence on external assistance. During continuous execution in the room, the average number of tasks completed autonomously increases from \textbf{0/10} to \textbf{6/10}. These results demonstrate how Zetta-RSR translates autonomously collected experience into more reliable and generalizable manipulation capabilities, enabling robots to complete longer task sequences with less human involvement and fewer agent takeovers.

% Faithful English translation of the approved Chinese Related Work, 2026-11-12.
% Three paragraphs and 15 sentences correspond to related_work_zh.md.
\section{Related work}
\label{sec:related-work}

\subsection{Agent--VLA collaboration and skill interfaces}

Robot agents organize low-level skills through high-level reasoning, providing a foundation for handling complex instructions and execution deviations.
SayCan~\cite{ahn2022can}, Voyager~\cite{wang2023voyager}, Eureka~\cite{ma2024eureka} and Code as Policies~\cite{liang2023code, fu2026cap} connect language reasoning to skill affordances and programmatic control, respectively, while Hi Robot~\cite{shi2025hi} further combines high-level task reasoning with a low-level VLA.
To improve execution reliability, Zetta~\cite{ding2026zetta} introduces runtime critics and recovery mechanisms, and Harness VLA~\cite{zhang2026harness} uses memory to coordinate a frozen VLA with analytic control; BATON~\cite{xu2026don} explicitly handles invocation readiness, skill handoffs, and constraints on successor states, while RoboAware~\cite{zhou2026roboaware} learns state-conditioned responsibility allocation from counterfactual execution outcomes.
These works provide a foundation for representing and learning collaboration conditions.
In this work, the Skill Interface describes the conditions under which the current VLA executes reliably, together with the agent's takeover, recovery, and control-return rules, thereby identifying operating states that require assistance and linking the reasons for intervention to the actual corrective process.

\subsection{From execution experience to internalized policy capabilities}

Autonomous data collection allows robots to accumulate experience for improving their own capabilities while executing tasks.
RoboCat~\cite{bousmalis2023robocat} augments its training set with experience generated autonomously by policies, while AutoRT~\cite{ahn2024autort} uses foundation models to organize real-robot data collection and broaden interaction coverage.
Recent work further integrates execution, verification, and training into automated closed loops: ENPIRE~\cite{xiao2026enpire} uses agents to manage real-world resets and policy improvement, HarnessPAI~\cite{wang2026harnesspai} and EmbodiedSWE~\cite{shen2026embodiedswe} turn trajectories produced by program execution into supervision for action models, and R2SI~\cite{manifold2026r2si} jointly updates skill programs and action models from the same execution records and revises invocation guidance as the model changes.
Building on these studies, we collect verified failure judgments and actual recovery trajectories from agent takeovers to train the VLA to autonomously recognize and handle the corresponding states.
The goal is to progressively turn operations that previously required agent assistance into capabilities of the policy itself, expanding the range of states the robot can handle autonomously~\cite{fang2026find,gupta2022bootstrapped}.

\subsection{Simulation-assisted continual improvement across domains}

Simulation provides repeatable trials for developing robot capabilities.
Visual domain randomization~\cite{tobin2017domain} and dynamics domain randomization~\cite{peng2018sim} promote model transfer by varying training conditions; recent methods further investigate cross-domain reuse of agent programs and experience: ASPIRE~\cite{lu2026aspire} applies verified repair knowledge to real-robot programming, EmbodiedRSI~\cite{song2026embodiedrsi} transfers a simulation-evolved harness to a real-robot system with a frozen VLA, and SimEX~\cite{hu2026simex} jointly revises the toolbox and simulator through a small number of real-world trials.
Recova~\cite{liu2026recova} further develops task and recovery behaviors in a digital twin and uses real-execution data to update the task and recovery policies separately.
In this work, a Skill Interface based on conditions observable in both domains connects simulation exploration, limited real-world adaptation, agent takeover, and learning from recovery experience; after each policy update, we revise the interface and organize the next round of data collection around cases that still require agent takeover.
The goal is to continually turn the judgment and corrective capabilities exercised by large models during interaction into autonomous capabilities of the robot policy, forming an iterative loop toward lifelong learning~\cite{wang2026self, sharma2023self} and recursive self-improvement (RSI)~\cite{han2026recursive, wang2026adarsi}.

\section{Method}
\label{sec:method}

\subsection{Problem Formulation}
\begin{figure*}[t]
  \centering
  \includegraphics[width=\textwidth]{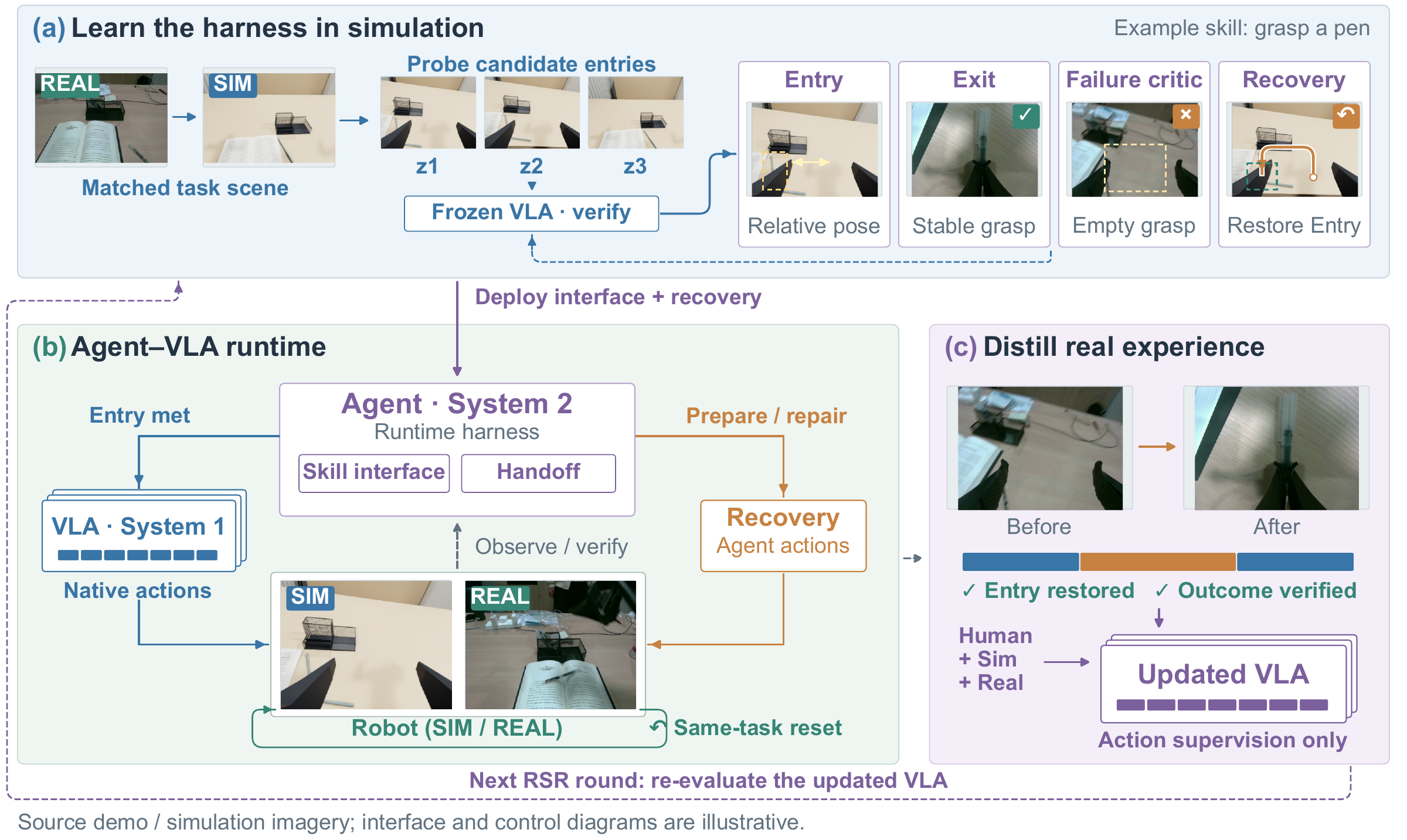}
  \caption{\textbf{Zetta-RSR framework.}
  (a) With the current VLA frozen, simulation trials inform per-skill Entry, Exit, Failure Critic, and Recovery components.
  (b) The agent and runtime harness coordinate native VLA actions with preparation or recovery, using observations from the same robot; Reset restarts the same task.
  (c) Action segments with verified recovery and subsequent outcomes join human and simulation data for post-training. The updated VLA is re-evaluated in the next round.
  \emph{Archived images and the projected correction path illustrate the method, rather than measured recovery results.}}
  \label{fig:framework}
\end{figure*}

We formalize data collection problem as repeated, automated task execution with Agent-VLA collaboration in both simulation and real. In each RSR round, we let an agent explore the VLA capability in simulation, apply learned Agent-VLA collaboration in both simulation and real, and collect verified trajectories to update the VLA. (Figure~\ref{fig:framework})

Given a task instruction $g$ and observations $o_t$, the VLA at RSR round $r$ outputs action chunks:
\begin{equation}
a_t=\pi_{\theta_r}(o_{\leq t},g).
\end{equation}
We initialize from a VLA $\pi_{\theta_0}$ post-trained on a small set of human demonstrations $\mathcal D_{\mathrm H}$. Due to the limited coverage and of initial human demonstration, VLA may exhibit inability to work for certain OOD layouts, instability in fine operation, or constant failure in certain task stages. Therefore, we divide tasks into minimal contact-rich manipulation skills $\{s_1,...,s_K\}$, and utilize an agent to assist the VLA through an interface including four components:
\begin{equation}
\mathcal I_{s}
=
(E_{s},X_{s},C_{s},R_{s}).
\end{equation}
Where entry $E_{s}$ describes the relative initial gripper-to-object poses in which the VLA can safely execute the skill $s$. Exit $X_{s}$ describes the observable outcome that occur after the skill is completed. Failure Critic $C_{s}$ detects failure risks or recoverable failures, and invoke recovery $R_{s}$ to mitigate the risk or restore a valid re-entry~\cite{ding2026zetta}. All interfaces use manifests that can be directly obtained or estimated in simulation and the real world for detection and control.

At each RSR round, the agent learns the VLA capability and interface, and accumulates verified  verified valid trajectories into the post-training dataset:
\begin{equation}
\mathcal D_{r}
=
\mathcal D_{\mathrm H}
\cup
\bigcup_{i=1}^{r}
\left(
\widetilde{\mathcal D}_{\mathrm{Sim}}^i
\cup
\widetilde{\mathcal D}_{\mathrm{Real}}^i
\right).
\end{equation}
Here, $\widetilde{\mathcal D}_{\mathrm{Sim}}^i$ and $\widetilde{\mathcal D}_{\mathrm{Real}}^i$ denote the filtered simulation trajectories and real-robot trajectories verified by their outcomes at round $i$, respectively. The VLA is post-trained on $\mathcal D_{r}$ to independently handle more states that previously required agent assistance.

\subsection{Learning Capability-Conditioned Skill Interfaces}

\noindent\textbf{Learning skill interfaces in simulation (Figure~\ref{fig:framework}a).}
we apply controlled perturbations to candidate entry states for each skill and execute the current
VLA. The agent looks for failure modes such as failure in certain OOD states, instability risk in fine operation, and systematic incapability in certain skills. Based on these findings, it updates the Interface $I_s$ and write controllers to intervene when the VLA runs into a failure mode. The agent may choose by itself to:
\begin{itemize}
    \item \textbf{Adjust before entry}: write a controller to modify entry position before handing to VLA through the $E_s$ interface.
    \item \textbf{Interrupt upon risk}: write a controller to stop VLA and restore to a safe state when detecting failure risk through the $C_s$ and $R_s$ interface.
    \item \textbf{Takeover if incapable}: If VLA is consistently incapable of performing a skill, takeover it with a controller when entering the skill through the $X_s$ interface.
\end{itemize}

\noindent\textbf{Runtime handoff and recovery (Figure~\ref{fig:framework}b).}
At runtime. The VLA executes the skills progressively. Upon entering contact-rich region, The agent checks entry condition and adjust the gripper if the entry condition is not met. Once the VLA completes the current skill, the exit
condition provides outcome evidence, and it enters the next skill.

When the critic identifies a failure risk or a recoverable deviation, the agent executes the corresponding recovery controller and restores VLA for retrial.

In the real world, the numerical intervention rules cannot be directly used due to the sensor imperfections, yet the failure modes of the VLA mostly hold. So we adapt the rules to real-world perception through a small amount of exploration, and deploy the interface in the real world~\cite{liu2026recova}. 
\subsection{Verified Real-world Distillation}
To internalize the agent interventions into the VLA's autonomous capabilities, we accumulate verified trajectories from both simulation and real-world for post-train. Simulation trajectories expose the VLA to more diverse layouts and agentic interventions, while real-world trajectories keep the VLA grounded to real-world perceptions. The VLA is post-trained with loss:
\begin{equation}
\mathcal L_{\mathrm{VLA}}(\theta_r;\mathcal D_{r})
=
\sum_{(o_{\leq t},g,a_t)\in\mathcal D_{r}}
-\log \pi_{\theta_r}(a_t\mid o_{\leq t},g).
\end{equation}

The update changes VLA capability boundaries, so in the next round of RSR, the agent modifies the interface and hands mastered abilities back to VLA. In this way, the agent’s external capabilities are progressively distilled into the VLA’s autonomous action capability. 
% Experiments section only; include from the main manuscript with \input{experiments}.
% Uses the main project preamble, bibliography, and existing figures/ directory.
% Sim-to-real interface transfer remains commented out.

\section{Experiments}
\label{sec:experiments}

Our experiments address three questions: (1) Can Zetta-RSR support continuous real-world operation and autonomous data collection with less human effort? (2) Can autonomously collected experience translate into sustained capability growth and improved generalization through multiple RSR rounds? (3) Can the agent's judgment and corrective behaviors be internalized by the VLA, and where does the key experience supporting this process come from?

\begin{figure*}[!t]
  \centering
  \includegraphics[width=\textwidth]{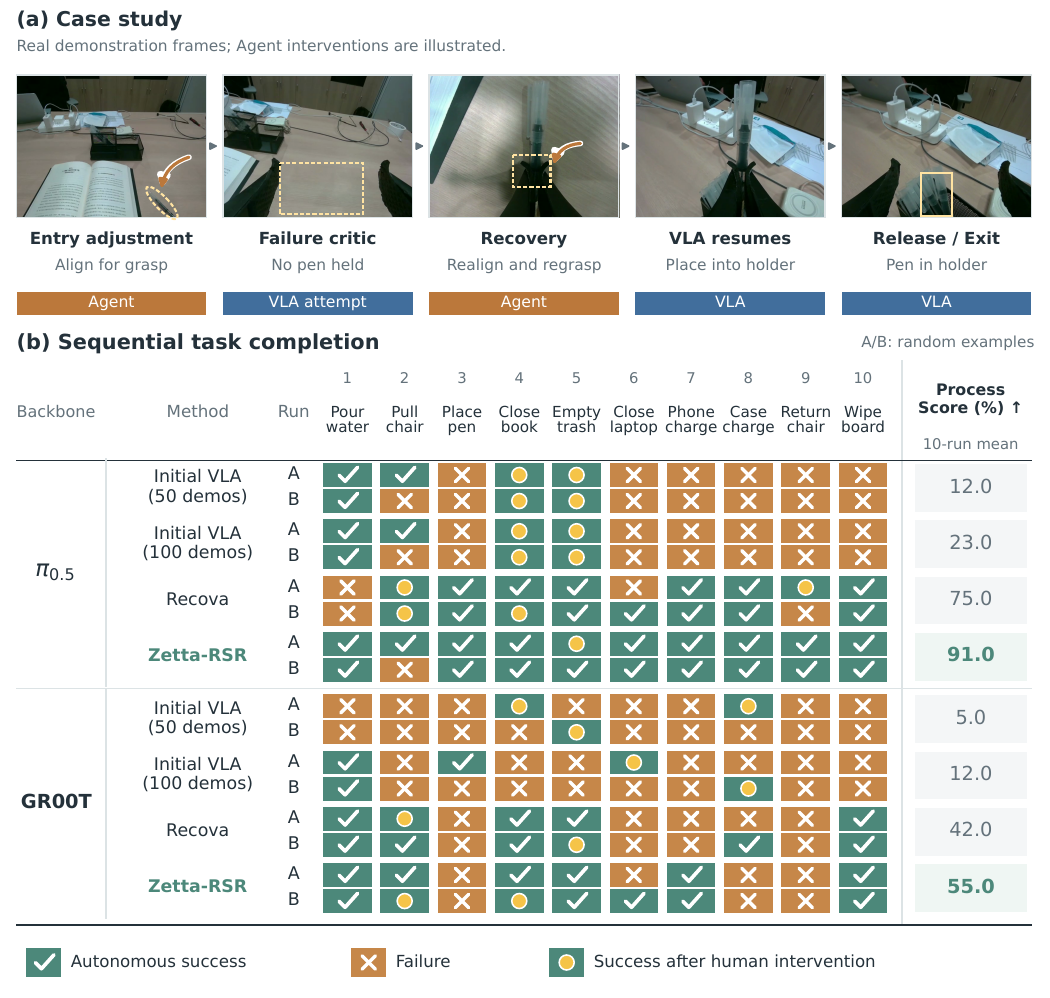}
  \caption{\textbf{Pen-placement recovery and sequential task completion.} (a) Real demonstration frames illustrate Entry adjustment, an empty-gripper check after a grasp attempt, Agent realignment and regrasp, and VLA placement and release. Orange/blue bars indicate illustrative Agent/VLA control; outlined regions and arrows highlight checked states and proposed adjustments. (b) Columns follow the ten-task order. Each configuration is evaluated in ten runs; A/B visualize two randomly selected runs, and the rightmost column reports the ten-run mean Process Score. Process Score (\%) averages the percentage of autonomously completed tasks over all ten runs; human-assisted tasks receive no credit, while later autonomous successes still count. Initial VLA variants use 50 or 100 demonstrations. Both Recova and Zetta-RSR are initialized from the corresponding backbone trained on 50 demonstrations. Green checks denote autonomous success, orange crosses denote failure, and green cells with yellow dots denote human-assisted success.}
  \label{fig:exp-room}
\end{figure*}

\begin{table*}[!t]
  \centering
  \caption{Five-task performance and data-collection costs. Each task cell and the mean column report system success rate/OOD success rate (\%). Training trajectories are cumulative across five tasks: 300 human demonstrations plus 600 simulation and 600 real-robot trajectories for each post-training method. Demo-equivalent hours normalize data volume by the mean human-demonstration duration (0.04 h/trajectory), rather than labor time. Human hours include initial collection and are reported approximately; timing and accounting details are given in Appendix~\ref{app:exp-cost}.}
  \label{tab:exp-main}
  \begingroup
  \small
  \setlength{\tabcolsep}{3pt}
  \resizebox{\textwidth}{!}{%
  \begin{tabular}{@{}llccccccccc@{}}
    \toprule
    Backbone & Method & \shortstack{Pick\\eraser $\uparrow$} & \shortstack{Clean\\whiteboard $\uparrow$} & \shortstack{Place\\pen $\uparrow$} & \shortstack{Throw\\rubbish $\uparrow$} & \shortstack{Charge\\earphone $\uparrow$} & \shortstack{Five-task\\mean $\uparrow$} & \shortstack{Training trajectories\\(count) $\uparrow$} & \shortstack{Demo-equivalent\\hours $\uparrow$} & \shortstack{Actual human\\hours $\downarrow$} \\
    \midrule
    $\pi_{0.5}$ & Initial VLA (300) & 36/18 & 48/26 & 40/6 & 40/26 & 0/0 & 32.8/15.2 & 300 & 12 & 32 \\
     $\pi_{0.5}$ & Initial VLA (600) & 72/56 & 76/58 & 78/30 & 60/50 & 56/40 & 68.4/46.8 & 600 & 24 & 64 \\
    $\pi_{0.5}$ & Recova adaptation & 48/34 & 64/44 & 66/56 & 94/84 & 46/20 & 63.6/47.6 & 1500 & 60 & $\approx 39.5$ \\
    $\pi_{0.5}$ & Zetta-RSR & 60/42 & 82/66 & 92/64 & 96/86 & 84/76 & 82.8/66.8 & 1500 & 60 & $\approx 38$ \\
    \midrule
    GR00T & Initial VLA (300) & 0/0 & 32/22 & 6/0 & 8/0 & 0/0 & 9.2/4.4 & 300 & 12 & 32 \\
        GR00T & Initial VLA (600) & 2/2 & 36/26 & 50/22 & 30/8 & 4/0 & 24.4/11.6 & 600 & 24 & 64 \\
    GR00T & Recova adaptation & 32/16 & 56/32 & 52/30 & 40/28 & 66/52 & 49.2/31.6 & 1500 & 60 & $\approx 40.5$ \\
    GR00T & Zetta-RSR & 50/20 & 66/40 & 72/58 & 34/26 & 66/48 & 57.6/38.4 & 1500 & 60 & $\approx 39$ \\
    \bottomrule
  \end{tabular}%
  }
  \endgroup
\end{table*}

\begin{figure*}[!t]
  \centering
  \includegraphics[width=\textwidth]{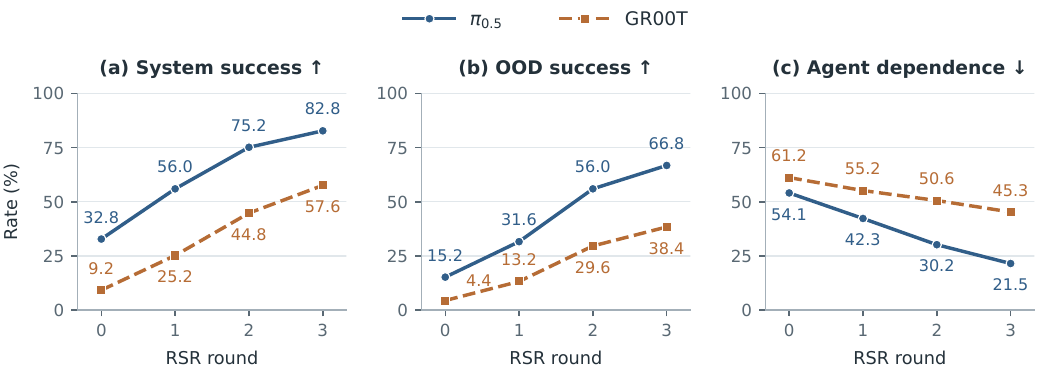}
  \caption{Self-improvement through multiple RSR rounds on $\pi_{0.5}$~\cite{intelligence2025pi_} and GR00T~\cite{bjorck2025gr00t} (a) System success rate. (b) OOD success rate. (c) Dependence on the Agent, measured as the percentage of attempts requiring agent takeover. Round 0 denotes the initial checkpoint, followed by three RSR updates. Success rates increase while dependence on the Agent decreases for both backbones.}
  \label{fig:exp-rsr}
\vspace{5mm}
\end{figure*}

\subsection{Experimental setup}
\label{sec:exp-setup}

We use an ARX Lift robot and cover ten everyday tasks: closing a laptop, placing a pen, closing a book, wirelessly charging a phone, pouring water and picking up the cup, charging an earbud case, disposing of trash, pushing a chair, pulling a chair, and wiping a blackboard. Table~\ref{tab:exp-main} reports a detailed evaluation on five of these tasks. For each of the $\pi_{0.5}$~\cite{intelligence2025pi_} and GR00T~\cite{bjorck2025gr00t} backbones, we compare the initial VLA, an adaptation of Recova's mechanism~\cite{liu2026recova}, and Zetta-RSR. All methods share mobile-base control and basic task scheduling. Within each backbone, the initial VLA is evaluated with 300 or 600 human demonstrations in Table~\ref{tab:exp-main}; Recova and Zetta-RSR start from the 300-demonstration checkpoint and use matched training-data counts and test conditions. The Recova adaptation retains the complete recovery pathway. Model configurations and task protocols are provided in Appendices~\ref{app:exp-platform}--\ref{app:exp-tasks}.

Our main metrics are system success rate, OOD success rate, training-trajectory count, demo-equivalent hours, and actual human person-hours. OOD tests cover unseen positions/orientations and lighting/backgrounds, with all configurations held out from training and interface tuning. The main table reports macro-averaged performance, cumulative training data, and recorded human effort across the five tasks, including initial demonstration collection. Metric definitions and functional success criteria are provided in Appendices~\ref{app:exp-tasks}--\ref{app:exp-cost}; detailed time measurement and person-hour accounting are given in Section~C.

\subsection{Main results}
\label{sec:exp-main}

\noindent\textbf{Room-scale continuous execution.}
We evaluate each backbone--method configuration in ten runs of the fixed ten-task sequence. Process Score (\%) is the percentage of the ten tasks completed autonomously in each run, averaged over all ten runs. Human-assisted tasks receive no credit; tasks completed autonomously after an earlier failure still count. Figure~\ref{fig:exp-room}(b) visualizes two runs (A/B) randomly selected from the ten runs of each configuration; these examples illustrate task completion and human intervention within a sequence, while quantitative scores use all ten runs. Zetta-RSR obtains Process Scores of \textbf{91\%} and \textbf{55\%} on $\pi_{0.5}$ and GR00T, respectively. Both Recova and Zetta-RSR start from the corresponding backbone trained on 50 demonstrations. Complete statistics are provided in Appendix~\ref{app:exp-room}.

\noindent\textbf{Pen-placement case study.}
Figure~\ref{fig:exp-room}(a) uses real demonstration frames to illustrate the collaboration mechanism for Entry adjustment and grasp recovery. During sequential task execution, the final state of one task may not satisfy the entry conditions of the next. The Agent first adjusts the grasping pose to satisfy Entry. During execution, if the gripper remains empty after a VLA grasp attempt, the Failure critic triggers Recovery based on the unmet object-holding condition. The Agent then realigns and regrasps, returning control to the VLA once the pen-holding state has been verified. The VLA continues with placement and release, and a final Exit check confirms that the pen is in the holder.

\noindent\textbf{Autonomous data collection and human costs.}
Table~\ref{tab:exp-main} shows the effectiveness of Zetta-RSR in expanding training data and reducing human data-collection effort. Starting from 300 human demonstrations across the five tasks, the system collects an additional 600 simulation trajectories and 600 real-robot trajectories, expanding the training set to 1,500 trajectories. Accounting for initial human collection and real-robot supervision, the cumulative human effort is approximately 38 and 39 person-hours on $\pi_{0.5}$ and GR00T, respectively. Compared with the 64 person-hours required to collect 600 human demonstrations, Zetta-RSR expands the training-data volume to 2.5 times as much while reducing recorded human effort by approximately 40.6\% and 39.1\%, respectively.

Under the same setup of four ARX Lift robots supervised by one operator, Zetta-RSR takes approximately 6 and 7 hours to collect 600 real-robot trajectories on the $\pi_{0.5}$ and  GR00T backbones respectively, shortening collection time by approximately 20.0\% and 17.6\% relative to Recova's 7.5 and 8.5 hours. The corresponding five-task mean system/OOD success rates are \textbf{82.8\%/66.8\%} on $\pi_{0.5}$ and \textbf{57.6\%/38.4\%} on GR00T, further assessing the contribution of the collected data in improving the task execution robustness and generalization ability. Stage-wise time measurements, the scope of human-effort accounting, and the data-source breakdown are provided in Appendix~\ref{app:exp-cost}.

\subsection{Self-improvement through multiple RSR rounds}
\label{sec:exp-rsr}

Figure~\ref{fig:exp-rsr} examines whether autonomous data collection and training can continuously improve system capabilities over multiple RSR rounds. Across the three updates, system/OOD success rates rise from 32.8\%/15.2\% to 82.8\%/66.8\% on $\pi_{0.5}$, and from 9.2\%/4.4\% to 57.6\%/38.4\% on GR00T. Over the same checkpoints, the agent takeover rate decreases from 54.1\% to 21.5\% and from 61.2\% to 45.3\%, respectively. Both backbones therefore improve task reliability and generalization while requiring agent preparation, adjustment, or recovery in fewer attempts. Evaluation protocols for each round are provided in Appendix~\ref{app:exp-analysis}.

\begin{figure}[!t]
  \centering
  \includegraphics[width=\linewidth]{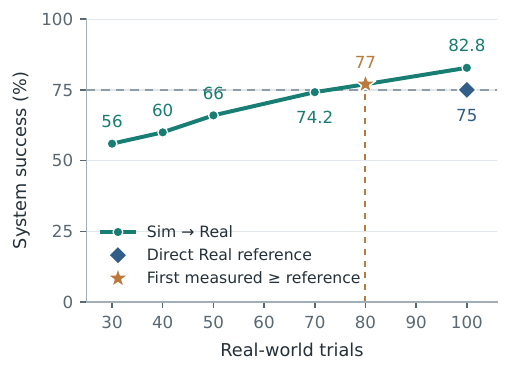}
  \caption{\textbf{Sim-to-real interface transfer.} System success versus real-world trials with the VLA fixed. Direct Real reaches 75\% after 100 trials; the star marks the first evaluated Sim$\rightarrow$Real checkpoint exceeding this reference (80 trials, 77\%).}
  \label{fig:exp-transfer}
\end{figure}

\subsection{Ablations and analysis}
\label{sec:exp-ablations}

\noindent\textbf{Distilling C\&R experience.}
Table~\ref{tab:exp-cr} compares the original VLA, the original VLA with external C\&R, and the C\&R-distilled VLA evaluated with external C\&R disabled. Adding external C\&R improves the initial VLA's system and OOD success rates from 32.8\%/15.2\% to 86.0\%/71.2\%. After distillation, the VLA achieves 79.2\%/64.4\% without runtime C\&R, improving upon the unassisted initial model by 46.4 and 49.2 percentage points, respectively. Its system and OOD success rates are each 6.8 percentage points below the externally assisted reference. These results support transferring agent-assisted execution experience into the VLA's action capabilities, improving task execution and generalization without runtime C\&R while leaving a performance gap to the assisted reference. The evaluation metrics are defined in Appendix~\ref{app:exp-tasks}.

\begin{table}[!t]
  \centering
  \caption{Joint C\&R ablation. External C\&R indicates whether external critic and recovery assistance is enabled. SR and OOD SR denote system success rate and OOD success rate, respectively. All values are in \%.}
  \label{tab:exp-cr}
  \begingroup
  \small
  \setlength{\tabcolsep}{3pt}
  \resizebox{\columnwidth}{!}{%
  \begin{tabular}{@{}lccc@{}}
    \toprule
    Configuration & \shortstack{External\\C\&R} & SR $\uparrow$ & OOD SR $\uparrow$ \\
    \midrule
    Original VLA       & No  & 32.8 & 15.2 \\
    Original VLA + C\&R & Yes & 86.0 & 71.2 \\
    C\&R-distilled VLA  & No & 79.2 & 64.4 \\
    \bottomrule
  \end{tabular}%
  }
  \endgroup
\end{table}

\noindent\textbf{Data sources.}
Table~\ref{tab:exp-data} shows that both simulation data and agent-collected real-robot data supplement the initial human demonstrations, with their combination achieving the highest system and OOD success rates on both backbones. Adding agent-collected real-robot data to human demonstrations and simulation data improves system/OOD success rates by 37.6/38.0 percentage points on $\pi_{0.5}$ and 16.4/12.0 percentage points on GR00T, demonstrating the contribution of real interaction experience to post-training. Compared with using only human demonstrations and agent-collected real-robot data, the full combination further improves OOD success by 4.8 and 12.0 percentage points, respectively, indicating that simulation trajectories provide additional generalization gains even when real-robot data are available. These results support jointly using data generated through simulation exploration and real-robot interaction in the RSR loop to continually expand the model's execution capabilities and generalization. Data splits and control conditions are described in Appendix~\ref{app:exp-analysis}.

\begin{table}[!t]
  \centering
  \caption{Data-source ablation on $\pi_{0.5}$ and GR00T, reported separately. Within each backbone, all configurations use the same system configuration and report system success rate and OOD success rate (\%).}
  \label{tab:exp-data}
  \begingroup
  \small
  \setlength{\tabcolsep}{4pt}
  \resizebox{\columnwidth}{!}{%
  \begin{tabular}{@{}ccccc@{}}
    \toprule
    \shortstack{Human\\$D_H$} & \shortstack{Sim\\$D_S$} & \shortstack{Agent real\\$D_A$} & \shortstack{System\\success $\uparrow$} & \shortstack{OOD\\success $\uparrow$} \\
    \midrule
    \multicolumn{5}{c}{\textbf{$\pi_{0.5}$}} \\
    \midrule
    $\checkmark$ & ---           & ---           & 32.8 & 15.2 \\
    $\checkmark$ & $\checkmark$ & ---           & 45.2 & 28.8 \\
    $\checkmark$ & $\checkmark$ & $\checkmark$ & 82.8 & 66.8 \\
    $\checkmark$ & ---           & $\checkmark$ & 80.0 & 62.0 \\
    \midrule
    \multicolumn{5}{c}{\textbf{GR00T}} \\
    \midrule
    $\checkmark$ & ---           & ---           & 9.2 & 4.4 \\
    $\checkmark$ & $\checkmark$ & ---           & 41.2 & 26.4 \\
    $\checkmark$ & $\checkmark$ & $\checkmark$ & 57.6 & 38.4 \\
    $\checkmark$ & ---           & $\checkmark$ & 49.6 & 26.4 \\
    \bottomrule
  \end{tabular}%
  }
  \endgroup
\end{table}

\noindent\textbf{Sim-to-real interface transfer.}
Figure~\ref{fig:exp-transfer} compares the real-world interaction efficiency of two interface-learning routes with the VLA weights held fixed. Learning the Interface directly in the real world achieves a success rate of 75\% after 100 trials. In contrast, Sim$\rightarrow$Real reaches 77\% after 80 trials, the first evaluated checkpoint exceeding this reference, using 20\% fewer real-world trials than the reference route. With the same budget of 100 real-world trials, Sim$\rightarrow$Real further reaches 82.8\% success, exceeding direct real-world learning by 7.8 percentage points. These results indicate that an Interface learned in simulation provides an effective starting point for real-world exploration, exceeding the reference performance with fewer real interactions and achieving higher system success under the same real-world interaction budget. The evaluation protocol is detailed in Appendix~\ref{app:exp-analysis}.

\vspace{2mm}
\section{Conclusion}
\vspace{2mm}
\label{sec:conclusion}
We proposed Zetta-RSR, a self-evolving autonomous data collection pipeline based on an Agent-VLA interface. We explore the interface in simulation, allowing the agent to learn the capability boundary of the VLA, and invoke recovery when the VLA falls short. We apply the interface in simulation and real world, collecting both scene-diverse and high-quality robot data. We progressively distill the agent assisting behavior into the VLA autonomous capability, expanding its capability region and improving its action robustness. We validated our pipeline on multiple real-world indoor tasks, achieving autonomous learning for a long-horizon household task series.
% \vspace{5mm}
% \section*{LLM Usage Statement}
% We used OpenAI ChatGPT and Codex to assist with editing manuscript text, writing code, and improving figures and diagrams. The research ideas and direction, the experimental design, and the analysis and interpretation of the results were developed and carried out by the human authors.

% Our work focuses on an automatic data-collection pipeline for VLA. All data except the initial human-curated data are produced by the agent and VLA on simulated and real-world robots. The human authors reviewed, cleaned but did not manually create the data trajectories.

\clearpage
\section*{LLM Usage Statement}
We used OpenAI ChatGPT and Codex to assist with editing manuscript text, writing code, and improving figures and diagrams. The research ideas and direction, the experimental design, and the analysis and interpretation of the results were developed and carried out by the human authors.

Our work focuses on an automatic data-collection pipeline for VLA. All data except the initial human-curated data are produced by the agent and VLA on simulated and real-world robots. The human authors reviewed, cleaned but did not manually create the data trajectories.

{\small
\bibliographystyle{ieeenat_fullname}
\bibliography{references}
}

\clearpage
\appendix
% Appendix content, included by main.tex after the references.
% This is not a standalone document.
\section*{Appendix}
\setcounter{figure}{0}
\setcounter{equation}{0}
\renewcommand{\thefigure}{S\arabic{figure}}
\renewcommand{\theequation}{S\arabic{equation}}
\renewcommand{\theHfigure}{appendix.\arabic{figure}}
\renewcommand{\theHtable}{appendix.\arabic{table}}
\renewcommand{\theHequation}{appendix.\arabic{equation}}
\raggedbottom
% Organized supplementary material, 2026-11-23.
% Standalone sections A--E retain the main paper's section-letter pointers.
% Unknown implementation details are tracked separately, not filled by inference.
\setcounter{table}{0}
\renewcommand{\thetable}{S\arabic{table}}

\noindent\textbf{Overview.}
Section~\ref{app:exp-platform} describes the platform, model roles, and initialization. Section~\ref{app:exp-tasks} defines the tasks and evaluation metrics. Section~\ref{app:exp-cost} details data composition and human-effort accounting. Section~\ref{app:exp-analysis} provides the RSR, ablation, and interface-transfer results. Section~\ref{app:exp-room} specifies sequential evaluation and Process Score. References to figures and tables in the main paper use their main-paper numbers; appendix tables and equations use S-prefixed numbering.

\section{Platform, models, and initial data}
\label{app:exp-platform}

\subsection{Robot and Agent--VLA roles}
Experiments use the ARX Lift platform with $\pi_{0.5}$ and GR00T as VLA backbones. Within each backbone, tasks share a model. The Agent controls the mobile base and coordinates manipulation; the VLA produces action chunks for manipulation. Mobile-base motion is part of the room execution, but navigation is not evaluated separately. Real-world data collection for the five-task comparison uses four ARX Lift robots supervised by one operator.

The skill interface comprises Entry, Exit, Failure Critic, and Recovery. Entry specifies conditions for invoking a VLA skill, Exit describes its expected observable completion state, and the Failure Critic identifies execution problems requiring recovery. The Agent adjusts or repairs the state and checks the relevant Entry before returning control to the VLA. These runtime decisions belong to the Agent and interface; the VLA is not assumed to output critic decisions or explicit requests for help. Post-training uses verified action experience to improve the VLA's execution capabilities.

\subsection{Robot and training configs}
\label{app:exp-configs}
The robot controller runs at 20 Hz. VLA inputs are RGB images from the head and two wrist cameras, the task instruction, and a 17-dimensional state (12 arm joints, two grippers, head yaw/pitch, and lift height). The 16-dimensional action gives an absolute end-effector position, a WXYZ quaternion, and a gripper target for each arm, in its arm-base frame.

We initialize $\pi_{0.5}$ from OpenPI \texttt{pi05\_base} and GR00T from \texttt{nvidia/GR00T-N1.7-3B}. Both use full-parameter training on eight H100 GPUs, with AdamW, BF16 computation, cosine learning-rate decay, gradient clipping at 1.0, and seed 42. Each data source is split into 90\% training and 10\% validation by complete trajectory. State and action values are normalized using training data only.

For $\pi_{0.5}$, the global batch size is 64, the peak learning rate is $2.5\times10^{-5}$, and weight decay is $10^{-10}$. Images are resized to $224\times224$, and the action horizon is 50 steps. For GR00T, the global batch size is 128, the peak learning rates are $10^{-5}$ for the backbone and $10^{-4}$ for the remaining policy parameters, and weight decay is $10^{-5}$ (zero for bias and normalization parameters). Images are resized to $256\times256$, and the action horizon is 40 steps. For the human:simulation:real mixture of 1:2:2, training uses 70,000/35,000 updates and 1,000/1,750 warmup updates for $\pi_{0.5}$/GR00T. At inference, both execute 20 actions per chunk; $\pi_{0.5}$ uses 10 denoising steps and GR00T uses four.

\subsection{Initialization and comparison groups}
For the five-task comparison in main-paper Table 1, the initial VLA variants use 300 or 600 human demonstrations in total across the five tasks. Recova adaptation and Zetta-RSR start from the 300-demonstration checkpoint of the corresponding backbone. Each post-training configuration adds 600 simulation and 600 real-robot trajectories. The Recova comparison adapts its task/recovery mechanism, including recovery verification, to the common platform.

The sequential experiment uses separately labeled 50- and 100-demonstration initial checkpoints. Both Recova adaptation and Zetta-RSR start from the corresponding 50-demonstration checkpoint in that experiment.

\section{Tasks and evaluation protocols}
\label{app:exp-tasks}

\subsection{Task sets and completion criteria}
\label{app:exp-task-definitions}
% Main-paper Table 1 evaluates picking up an eraser, cleaning a whiteboard, placing a pen, disposing of rubbish, and charging an earbud case. The room sequence contains ten tasks in the fixed order shown in Table~\ref{tab:supp-tasks}. Picking up an eraser is an additional individual task and is not a separate item in this sequence. ``Charge earphone'' refers to placing the earbud charging case on a wireless charging stand; ``Clean whiteboard'' and ``Wipe blackboard'' refer to the board-cleaning task in the individual and sequential evaluations, respectively.
The full task sequence contains ten indoor tasks in the fixed order shown in Table~\ref{tab:supp-tasks}. The final task, Wipe Board, contains two separate individual tasks, pick eraser and clean whiteboard, making there 11 individual tasks in total. 5 among the 11 individual tasks are separately evaluated in Main-paper Table 1.

\begin{table*}[t]
\centering
\caption{Evaluation task sets and functional completion criteria. The sequence index gives the order in main-paper Figure 3. A checkmark indicates inclusion in the five-task comparison.}
\label{tab:supp-tasks}
\small
\setlength{\tabcolsep}{5pt}
\begin{tabularx}{\textwidth}{@{}clcX@{}}
\toprule
\shortstack{Sequence\\index} & Task & \shortstack{Table 1\\task} & Functional completion criterion \\
\midrule
1 & Pour water and pick up cup & --- & Complete pouring and retrieve the cup while satisfying the task's volume and spillage requirements. \\
2 & Pull chair & --- & Move the chair into the specified pull-out region. \\
3 & Place pen & $\checkmark$ & Place the pen in the specified holder and release it stably; grasping alone is insufficient. \\
4 & Close book & --- & Close the book and leave it stable. \\
5 & Dispose of rubbish & $\checkmark$ & Transfer the designated rubbish into the destination container. \\
6 & Close laptop & --- & Close the lid to the required terminal state. \\
7 & Charge phone & --- & Place the phone on the wireless charger so that charging starts. \\
8 & Charge earbud case & $\checkmark$ & Place the charging case on its wireless charging stand so that charging starts. \\
9 & Return chair & --- & Push the chair back into the specified placement region. \\
10 & Pick eraser & $\checkmark$ & Grasp and lift the board eraser, maintaining a stable hold. \\
10 & Clean whiteboard & $\checkmark$ & Clean the specified board area to the task's required condition. \\
\bottomrule
\end{tabularx}
\end{table*}

Task success is defined by the task outcome. Physical placement and functional outcomes are distinguished: for charging tasks, proximity to a charger alone does not satisfy the task goal. Human-assisted completion is not counted as autonomous success. Agent intervention remains part of autonomous system execution when it is included in the evaluated configuration.

\subsection{System success and OOD evaluation}
\label{app:exp-budgets}
The in-distribution evaluation uses conditions close to the initial-demonstration distribution. OOD evaluation targets unseen object positions/orientations and visual conditions such as illumination and background. These conditions are treated as held out from training and interface tuning.

\label{app:exp-aggregation}
For task $t$ and condition set $c\in\{\mathrm{ID},\mathrm{OOD}\}$, let $m_{t,c}$ and $n_{t,c}$ denote the numbers of successful and total evaluated attempts, and let $s_{t,c}=m_{t,c}/n_{t,c}$. The five-task means in Table 1 are macro-averages:
\begin{equation}
\mathrm{SR}=\frac{100}{5}\sum_{t=1}^{5}s_{t,\mathrm{ID}},\qquad
\mathrm{SR}_{\mathrm{OOD}}=\frac{100}{5}\sum_{t=1}^{5}s_{t,\mathrm{OOD}}.
\end{equation}
Thus, each task has equal weight in the reported mean. Each result refers to its stated deployment configuration, including whether external C\&R is enabled.

\subsection{Independent collection and Reset}
Data are collected independently for each task. After an attempt, Reset restores the scene for another attempt of the same task; the collection procedure does not require chaining different tasks. Reset concerns scene preparation, whereas Recovery addresses an execution problem within the ongoing attempt. Room-scale task chaining is reserved for the sequential evaluation in Section~\ref{app:exp-room}.

\subsection{Note on Real-World Evaluation}
\label{sec:eval_note}

Real-world VLA evaluation is subject to temporal variability
in evaluation conditions, including variations in manually
reset object poses, illumination, and hardware states.
To minimize the impact of such variability on comparative
results, we conduct all evaluations within each experimental
group during the same time window.

This also explains why the same $\pi_0.5$ baseline
($H:S:R=1:0:0$) appears with slightly different success
rates in the first and third experimental groups as reported in Tab.~\ref{tab:supp_full_real_results}, despite
using identical model checkpoints and inference settings.
Specifically, across 500 rollouts per evaluation window,
the overall success rates are 24.0\% and 26.0\%,
respectively, yielding a difference of 2.0 percentage
points (pp), with a 95\% confidence interval of
$[-2.97, +6.94]$ pp.

These discrepancies reflect potential inter-session
variability and finite-trial uncertainty.
Accordingly, all comparative improvements are primarily
assessed against the corresponding within-group baseline.

\section{Data composition and human-effort accounting}
\label{app:exp-cost}

\subsection{Trajectory counts}
\label{app:exp-valid-data}
The training-trajectory count in Table 1 is cumulative across the five tasks and includes all three data sources. Table~\ref{tab:supp-data-counts} separates initial demonstrations from additional simulation and real-robot data. Only accepted trajectories are counted as dataset entries, failured are not included.

\begin{table}[t]
\centering
\caption{Training-data composition for each backbone in main-paper Table 1. Counts are totals across the five tasks.}
\label{tab:supp-data-counts}
\small
\setlength{\tabcolsep}{4pt}
\begin{tabular}{@{}lrrrr@{}}
\toprule
Configuration & Human & Sim & Real & Total \\
\midrule
Initial VLA (300) & 300 & 0 & 0 & 300 \\
Initial VLA (600) & 600 & 0 & 0 & 600 \\
Recova adaptation & 300 & 600 & 600 & 1,500 \\
Zetta-RSR & 300 & 600 & 600 & 1,500 \\
\bottomrule
\end{tabular}
\end{table}

\subsection{Demo-equivalent hours}
\label{app:exp-human-cost}
The 600 human demonstrations contain 24 hours of recorded trajectory data. Their mean recorded duration is $\bar d_H=24/600=0.04$ hours per trajectory. We express training-data volume as
\begin{equation}
H_{\mathrm{demo}}=N_{\mathrm{train}}\bar d_H,
\end{equation}
where $N_{\mathrm{train}}$ includes human, simulation, and real-robot trajectories. The 300-, 600-, and 1,500-trajectory datasets therefore correspond to 12, 24, and 60 demo-equivalent hours. This is a data-volume normalization, not measured labor savings. In particular, 60 demo-equivalent hours do not mean 60 hours of real-robot recordings or equal training value across data sources.

\subsection{Actual human person-hours}
\noindent\textbf{Initial demonstrations.}
Two operators collected 600 demonstrations over four eight-hour working days, giving $2\times4\times8=64$ person-hours. This accounting includes preparation, Reset, and breaks during the collection shifts. The 24 hours of recorded trajectories are data duration, not operator working time. The 300-demonstration seed is assigned 32 person-hours in Table 1 and is charged once per configuration.

\noindent\textbf{Real-robot supervision.}
Four robots collect data under one operator. The recorded session durations for 600 real-robot trajectories are 7.5/8.5 hours for Recova on $\pi_{0.5}$/GR00T, and 6/7 hours for Zetta-RSR. All time assigned to supervision is counted, including periods without manual intervention. With one supervising operator, person-hours equal session duration; multiplying by four would measure allocated robot-hours instead.

The cumulative human effort reported in Table 1 is
\begin{equation}
H_{\mathrm{reported}}=H_{\mathrm{seed}}+n_{\mathrm{sup}}T_{\mathrm{real}},
\end{equation}
where $H_{\mathrm{seed}}=32$ person-hours and $n_{\mathrm{sup}}=1$. Table~\ref{tab:supp-human-costs} gives the calculation. Approximate signs reflect reported session-time estimates rather than exact per-action timing.

\begin{table}[t]
\centering
\caption{Recorded human effort for the post-training configurations, in person-hours. Each row includes the shared initial collection once.}
\label{tab:supp-human-costs}
\small
\setlength{\tabcolsep}{4pt}
\begin{tabular}{@{}llrrr@{}}
\toprule
Backbone & Method & Seed & Supervision & Total \\
\midrule
$\pi_{0.5}$ & Recova & 32 & 7.5 & $\approx39.5$ \\
$\pi_{0.5}$ & Zetta-RSR & 32 & 6.0 & $\approx38.0$ \\
GR00T & Recova & 32 & 8.5 & $\approx40.5$ \\
GR00T & Zetta-RSR & 32 & 7.0 & $\approx39.0$ \\
\bottomrule
\end{tabular}
\end{table}

\noindent\textbf{Simulation and accounting scope.}
Simulation collection runs unattended for five hours on eight RTX 5090 GPUs, corresponding to 40 allocated GPU-hours. Unattended runtime and GPU-hours are not human labor. Simulation preparation/debugging labor has not been logged and is excluded from the reported person-hours. The reported scope therefore comprises initial collection and recorded real-world supervision, rather than total system-development effort. Neither the 64-to-38/39 person-hour comparison nor the 600-to-1,500 trajectory comparison is a matched-volume comparison of real-robot data: the latter dataset also contains simulation trajectories.

\section{RSR iterations and analysis experiments}
\label{app:exp-analysis}

\subsection{Multiple RSR rounds}
\label{app:exp-rounds}
Main-paper Figure 4 reports the initial checkpoint, indexed by $r=0$, and three subsequent RSR updates. These four evaluated checkpoints do not fix the number of rounds required by the method. Table~\ref{tab:supp-rounds} provides the numerical values for both backbones. Success rates are evaluated at each checkpoint, rather than accumulated as ``at least one success'' across rounds.

Agent dependence is the percentage of task attempts requiring at least one Agent preparation, adjustment, or Recovery intervention. Repeated interventions within one attempt still contribute one to the numerator. Ordinary task invocation, mobile-base motion, and recovery performed directly by the VLA are excluded. This metric describes dependence on the Agent, not human intervention.

\begin{table}[t]
\centering
\caption{Values plotted in main-paper Figure 4. All rates are percentages. Agent dependence counts attempts requiring takeover.}
\label{tab:supp-rounds}
\small
\setlength{\tabcolsep}{5pt}
\begin{tabular}{@{}lrrrr@{}}
\toprule
Backbone & Round & SR & OOD SR & \shortstack{Agent\\dependence} \\
\midrule
$\pi_{0.5}$ & 0 & 32.8 & 15.2 & 54.1 \\
 & 1 & 56.0 & 31.6 & 42.3 \\
 & 2 & 75.2 & 56.0 & 30.2 \\
 & 3 & 82.8 & 66.8 & 21.5 \\
\midrule
GR00T & 0 & 9.2 & 4.4 & 61.2 \\
 & 1 & 25.2 & 13.2 & 55.2 \\
 & 2 & 44.8 & 29.6 & 50.6 \\
 & 3 & 57.6 & 38.4 & 45.3 \\
\bottomrule
\end{tabular}
\end{table}

\subsection{C\&R experience distillation}
\label{app:exp-cr}
Main-paper Table 2 evaluates the configurations in Table~\ref{tab:supp-cr}. External C\&R is enabled only for the assisted original VLA. The distilled VLA is evaluated with external C\&R disabled and achieves 79.2\% SR and 64.4\% OOD SR. Relative to the unassisted initial VLA, the gains are 46.4 and 49.2 percentage points; the gaps to the assisted reference are 6.8 percentage points for both metrics. These task-level scores assess execution after distillation without runtime C\&R; they do not separately quantify critic prediction accuracy or recovery success from individual failure states.

\begin{table}[t]
\centering
\caption{C\&R configurations and results from main-paper Table 2. Values are percentages.}
\label{tab:supp-cr}
\small
\setlength{\tabcolsep}{4pt}
\begin{tabular}{@{}lcrr@{}}
\toprule
Configuration & \shortstack{External\\C\&R} & SR & OOD SR \\
\midrule
Original VLA & No & 32.8 & 15.2 \\
Original VLA + C\&R & Yes & 86.0 & 71.2 \\
C\&R-distilled VLA & No & 79.2 & 64.4 \\
\bottomrule
\end{tabular}
\end{table}

\subsection{Data-source ablation}
\label{app:exp-data}
The four data groups are $D_H$, $D_H+D_S$, $D_H+D_A$, and $D_H+D_S+D_A$, where $D_H$ denotes human demonstrations, $D_S$ simulation trajectories, and $D_A$ Agent-collected real-robot trajectories. The comparison is performed separately for each backbone. The deployment configuration retains the Agent, interface, and C\&R while changing the VLA trained on each data mixture. Table~\ref{tab:supp-data-ablation} reports the results. Excluding $D_S$ from VLA training does not imply that the collector producing $D_A$ was developed without simulation; this ablation concerns training-data sources, rather than an entirely simulation-free pipeline.

\begin{table}[t]
\centering
\caption{Data-source ablation from main-paper Table 3. Each cell gives SR/OOD SR (\%).}
\label{tab:supp-data-ablation}
\small
\setlength{\tabcolsep}{6pt}
\begin{tabular}{@{}lrr@{}}
\toprule
Training data & $\pi_{0.5}$ & GR00T \\
\midrule
$D_H$ & 32.8/15.2 & 9.2/4.4 \\
$D_H+D_S$ & 45.2/28.8 & 41.2/26.4 \\
$D_H+D_A$ & 70.0/62.0 & 49.6/26.4 \\
$D_H+D_S+D_A$ & 82.8/66.8 & 57.6/38.4 \\
\bottomrule
\end{tabular}
\end{table}

\subsection{Sim-to-real interface transfer}
\label{app:exp-transfer}
This comparison holds the VLA weights fixed. Direct Real learns the interface in the real environment and provides a reference of 75\% SR after 100 real-world trials. Sim$\rightarrow$Real learns the interface in simulation before real-world refinement. Its evaluated checkpoints are listed in Table~\ref{tab:supp-transfer} and plotted in main-paper Figure 5. Trial counts are development interactions, not the denominators of the success-rate evaluations.

\begin{table}[t]
\centering
\caption{Interface-transfer checkpoints. Direct Real provides the 100-trial reference; the Sim$\rightarrow$Real curve consists only of the six listed observations.}
\label{tab:supp-transfer}
\small
\setlength{\tabcolsep}{7pt}
\begin{tabular}{@{}lrr@{}}
\toprule
Route & Real-world trials & SR (\%) \\
\midrule
Direct Real & 100 & 75.0 \\
\midrule
Sim$\rightarrow$Real & 30 & 56.0 \\
 & 40 & 60.0 \\
 & 50 & 66.0 \\
 & 70 & 74.2 \\
 & 80 & 77.0 \\
 & 100 & 82.8 \\
\bottomrule
\end{tabular}
\end{table}

The first evaluated Sim$\rightarrow$Real checkpoint exceeding the reference is 80 trials at 77\% SR, using 20\% fewer trials than the 100-trial reference. At 100 trials, its 82.8\% SR is 7.8 percentage points higher. No threshold crossing is inferred between checkpoints. The comparison reports real-interaction counts and point estimates; it does not establish statistical equivalence or measure total development cost, including simulation preparation.

\section{Sequential execution and Process Score}
\label{app:exp-room}

\subsection{Ten-task sequence}
The sequence is: pour water and pick up the cup, pull the chair, place the pen, close the book, dispose of rubbish, close the laptop, charge the phone, charge the earbud case, return the chair, and clean the whiteboard. Each backbone--method configuration is evaluated in ten runs. Main-paper Figure 3 shows two randomly selected examples, A/B, from these runs; its Process Scores use all ten runs.

\subsection{Scoring}
Let $y_{ij}=1$ when the system autonomously completes task $j$ in run $i$, and $y_{ij}=0$ otherwise. For $N=10$ runs and $K=10$ tasks, the reported score is
\begin{equation}
\mathrm{Process\ Score}=\frac{100}{NK}\sum_{i=1}^{N}\sum_{j=1}^{K}y_{ij}.
\label{eq:process-score}
\end{equation}
Green checks indicate autonomous success; orange crosses indicate failure; green cells with yellow dots indicate human-assisted success. Human-assisted successes, failures, and unattempted tasks receive zero credit, and the denominator remains ten tasks per run. An earlier failure or human intervention does not invalidate later tasks completed autonomously. Agent assistance is part of autonomous execution when enabled for the evaluated system.

Process Score measures completion across the sequence. It is neither the length of the uninterrupted successful prefix nor the proportion of runs completing all ten tasks without human assistance. The aggregate scores in Table~\ref{tab:exp-room-full} cannot recover all-ten-task success rates, human-intervention counts, run durations, or variability across runs; those quantities require the underlying run records.

\begin{table}[t]
\centering
\caption{Reported Process Scores over ten runs per configuration. Initial checkpoint labels follow main-paper Figure 3; Recova and Zetta-RSR start from the corresponding 50-demonstration checkpoint.}
\label{tab:exp-room-full}
\small
\setlength{\tabcolsep}{4pt}
\begin{tabular}{@{}llrr@{}}
\toprule
Backbone & Configuration & Runs & \shortstack{Process\\Score (\%)} \\
\midrule
$\pi_{0.5}$ & Initial VLA (50) & 10 & 12 \\
 & Initial VLA (100) & 10 & 23 \\
 & Recova adaptation & 10 & 75 \\
 & Zetta-RSR & 10 & 91 \\
\midrule
GR00T & Initial VLA (50) & 10 & 5 \\
 & Initial VLA (100) & 10 & 12 \\
 & Recova adaptation & 10 & 42 \\
 & Zetta-RSR & 10 & 55 \\
\bottomrule
\end{tabular}
\end{table}

\subsection{Pen-placement case study}
Main-paper Figure 3(a) uses real demonstration frames to illustrate Entry adjustment, a grasp check, Agent realignment and regrasp, and VLA placement and release. The preceding task's terminal state may not meet the next skill's Entry conditions. The illustrated recovery addresses an empty gripper after a grasp attempt, followed by adjustment and regrasp before control returns to the VLA. The control bars and proposed adjustments explain the collaboration mechanism; they are not an additional quantitative run or evidence that the illustrated intervention occurred in one of the measured A/B executions.

% Full task-level results from Model_Task_Success_Rates2.numbers.
% Requires booktabs (already loaded in the supplementary entry point).
\begin{table*}[t]
    \centering
    \caption{Detailed real-world success rates (\%) across five tasks
    under Normal (N) and out-of-distribution (OOD) conditions.
    $H:S:R$ denotes the ratio of human-collected data, simulation
    rollouts, and real-world rollouts, respectively. The average
    success rates are computed with equal task weights; Overall
    further averages the Normal and OOD scores. Each experimental
    group was evaluated within its own time window. Bold indicates the
    highest overall SR within each group.}
    \label{tab:supp_full_real_results}
    \small
    \setlength{\tabcolsep}{2.8pt}
    \renewcommand{\arraystretch}{1.17}
    \begin{tabular*}{\textwidth}{@{\extracolsep{\fill}}l*{10}{r}*{3}{r}@{}}
        \toprule
        & \multicolumn{2}{c}{\shortstack{Pick\\Eraser}}
        & \multicolumn{2}{c}{\shortstack{Erase\\Whiteboard}}
        & \multicolumn{2}{c}{\shortstack{Pick\\Pen}}
        & \multicolumn{2}{c}{\shortstack{Dispose\\Trash}}
        & \multicolumn{2}{c}{\shortstack{Charge\\Earphones}}
        & \multicolumn{3}{c}{Average SR} \\
        \cmidrule(lr){2-3}\cmidrule(lr){4-5}
        \cmidrule(lr){6-7}\cmidrule(lr){8-9}
        \cmidrule(lr){10-11}\cmidrule(l){12-14}
        $H:S:R$ & N & OOD & N & OOD & N & OOD & N & OOD
                & N & OOD & N & OOD & Overall \\
        \midrule
        \multicolumn{14}{l}{\textbf{Group 1: $\boldsymbol{\pi}_{0.5}$}} \\
        1:0:0 & 36 & 18 & 48 & 26 & 40 & 6 & 40 & 26 & 0 & 0 & 32.8 & 15.2 & 24.0 \\
        1:1:0 & 42 & 22 & 48 & 26 & 50 & 8 & 54 & 7 & 36 & 10 & 46.0 & 14.6 & 30.3 \\
        1:1:1 & 42 & 26 & 58 & 32 & 54 & 12 & 88 & 74 & 38 & 14 & 56.0 & 31.6 & 43.8 \\
        1:2:1 & 50 & 28 & 44 & 40 & 64 & 42 & 90 & 78 & 52 & 46 & 60.0 & 46.8 & 53.4 \\
        1:2:2 & 64 & 46 & 74 & 56 & 70 & 32 & 98 & 90 & 70 & 56 & 75.2 & 56.0 & 65.6 \\
        1:3:2 & 64 & 50 & 72 & 54 & 82 & 48 & 98 & 88 & 82 & 70 & 79.6 & 62.0 & 70.8 \\
        1:3:3 & 60 & 42 & 82 & 66 & 92 & 64 & 96 & 86 & 84 & 76 & 82.8 & 66.8 & 74.8 \\
        1:3:0 & 64 & 40 & 64 & 48 & 34 & 24 & 54 & 28 & 10 & 4 & 45.2 & 28.8 & 37.0 \\
        1:0:3 & 76 & 62 & 70 & 54 & 88 & 62 & 90 & 74 & 76 & 58 & 80.0 & 62.0 & 71.0 \\
        1:5:5 & 62 & 48 & 82 & 72 & 90 & 64 & 96 & 88 & 88 & 80 & 83.6 & 70.4 & \textbf{77.0} \\
        2:0:0 & 72 & 56 & 79 & 58 & 78 & 30 & 60 & 50 & 56 & 40 & 69.0 & 46.8 & 57.9 \\
        \midrule
        \multicolumn{14}{l}{\textbf{Group 2: GR00T}} \\
        1:0:0 & 0 & 0 & 32 & 22 & 6 & 0 & 8 & 0 & 0 & 0 & 9.2 & 4.4 & 6.8 \\
        1:1:0 & 0 & 0 & 32 & 26 & 22 & 10 & 12 & 6 & 2 & 0 & 13.6 & 8.4 & 11.0 \\
        1:1:1 & 22 & 10 & 40 & 24 & 46 & 28 & 12 & 4 & 6 & 0 & 25.2 & 13.2 & 19.2 \\
        1:2:1 & 28 & 16 & 40 & 22 & 52 & 28 & 16 & 10 & 44 & 24 & 36.0 & 20.0 & 28.0 \\
        1:2:2 & 40 & 22 & 46 & 30 & 64 & 50 & 16 & 8 & 58 & 38 & 44.8 & 29.6 & 37.2 \\
        1:3:2 & 42 & 28 & 46 & 32 & 70 & 44 & 18 & 12 & 54 & 42 & 46.0 & 31.6 & 38.8 \\
        1:3:3 & 50 & 20 & 66 & 40 & 72 & 58 & 34 & 26 & 66 & 48 & 57.6 & 38.4 & 48.0 \\
        1:3:0 & 28 & 14 & 40 & 28 & 56 & 32 & 24 & 16 & 58 & 42 & 41.2 & 26.4 & 33.8 \\
        1:0:3 & 46 & 22 & 54 & 36 & 60 & 36 & 28 & 14 & 60 & 44 & 49.6 & 30.4 & 40.0 \\
        1:5:5 & 50 & 28 & 72 & 44 & 70 & 64 & 40 & 28 & 72 & 44 & 60.8 & 41.6 & \textbf{51.2} \\
        2:0:0 & 2 & 2 & 36 & 26 & 50 & 22 & 30 & 8 & 4 & 0 & 24.4 & 11.6 & 18.0 \\
        \midrule
        \multicolumn{14}{l}{\textbf{Group 3: $\boldsymbol{\pi}_{0.5}$ (Sim: Agent+VLA rollout)}} \\
        1:0:0 & 34 & 20 & 52 & 34 & 42 & 14 & 42 & 22 & 0 & 0 & 34.0 & 18.0 & 26.0 \\
        1:1:1 & 48 & 30 & 54 & 36 & 52 & 4 & 84 & 78 & 36 & 20 & 54.8 & 33.6 & 44.2 \\
        1:2:2 & 60 & 54 & 80 & 60 & 78 & 40 & 92 & 92 & 78 & 62 & 77.6 & 61.6 & 69.6 \\
        1:3:3 & 64 & 44 & 86 & 70 & 98 & 62 & 92 & 88 & 88 & 70 & 85.6 & 66.8 & 76.2 \\
        1:5:5 & 64 & 46 & 76 & 76 & 98 & 60 & 94 & 94 & 80 & 80 & 82.4 & 71.2 & \textbf{76.8} \\
        \bottomrule
    \end{tabular*}
\end{table*}

\end{document}